\documentclass[10pt,conference,a4paper]{IEEEconf_ICPR18/IEEEtran}

\usepackage[bookmarks=false]{hyperref}
\usepackage[utf8]{inputenc}
\usepackage{amsmath, amsthm, amssymb}
\usepackage{algorithmic}
\usepackage{mathtools}

\usepackage[noadjust]{cite}

\hypersetup{
    colorlinks=true,        
    linkcolor=black,        
    citecolor=black,        
    filecolor=black,      
    urlcolor=black           
}

\usepackage{tablefootnote}
\usepackage{subcaption}
\usepackage{color}

\usepackage{listings}

\usepackage{tikz}
\usepackage{pgfplots}
\usepgfplotslibrary[units]

\begin{document}

\title{DLL: A Blazing Fast Deep Neural Network Library}

\author{\IEEEauthorblockN{Baptiste Wicht\IEEEauthorrefmark{1}, Andreas Fischer\IEEEauthorrefmark{2}, Jean Hennebert\IEEEauthorrefmark{3}}
\IEEEauthorblockA{
    HES-SO, University of Applied Science of Western Switzerland\\
    University of Fribourg, Switzerland
}\\
Email: \IEEEauthorrefmark{1}baptiste.wicht@hefr.ch, \IEEEauthorrefmark{2}andreas.fischer@unifr.ch, \IEEEauthorrefmark{3}jean.hennebert@hefr.ch}

\maketitle

\begin{abstract}

    Deep Learning Library (DLL) is a new library for machine learning with deep
    neural networks that focuses on speed. It supports feed-forward neural
    networks such as fully-connected Artificial Neural Networks (ANNs) and
    Convolutional Neural Networks (CNNs).  It also has very comprehensive
    support for Restricted Boltzmann Machines (RBMs) and Convolutional RBMs.
    Our main motivation for this work was to propose and evaluate novel software
    engineering strategies with potential to accelerate runtime for training and
    inference. Such strategies are mostly independent of the underlying deep
    learning algorithms.  On three different datasets and for four different
    neural network models, we compared DLL to five popular deep learning
    frameworks.  Experimentally, it is shown that the proposed framework is
    systematically and significantly faster on CPU and GPU. In terms of
    classification performance, similar accuracies as the other frameworks are
    reported.

\end{abstract}

\section{Introduction}

In recent years, neural networks have regained a large deal of attention with
deep learning approaches. Such approaches rely on the use of bigger and deeper
networks, typically by using larger input dimensions to incorporate more context
and by increasing the number of layers to extract information at different
levels of granularity. The success of deep learning can be attributed mainly
to three factors. First, there is the advent of \emph{big data}, meaning the
availability of larger quantities of training data. Second, new training
strategies have been developed, such as unsupervised pre-training that allows
deep networks to initialize well and also to learn efficient feature extractors
on large sets of unlabelled data. Finally, better and faster hardware has
helped dealing with the training of such networks.  Deep systems are currently
improving the state-of-the-art in many domains.  Successful deep learning
applications include near-human performance at recognizing objects in
images~\cite{Szegedy2015}, generating detailed image
descriptions~\cite{Karpathy2015}, adding colors to grayscale
images~\cite{Cheng2015} or generating highly-realistic
images~\cite{Goodfellow2014}. Moreover, the availability of free and easy-to-use
frameworks, as well as the availability of detailed implementation examples on
public datasets, have contributed to the widespread use of deep learning
technologies.

From a practical point of view, an ideal deep learning framework would be easy
to use, would offer fast training with good precision and would be versatile
with many configuration options. Reaching all these qualities is difficult as
some are contradictory. For this reason, we may observe large differences among
the available frameworks.

In this work, we report on the development of a new deep learning framework
where we have clearly opted to focus on efficient computation, targeting
specific network models and algorithm configurations. While we are aware of
these limitations, we believe that the different optimizations we have
implemented in our framework may be of interest to the scientific community. Our
framework is called Deep Learning Library (DLL) and is freely available, with
source code\footnote{URL \url{https://github.com/wichtounet/dll}}. The initial
reason behind the development of a complete framework was the lack of Restricted
Boltzmann Machine (RBM)~\cite{Smolensky1986} and Convolutional RBM
(CRBM)~\cite{Lee2009} support in other Machine Learning frameworks. This is
still the case at the time of writing. Along the way, the framework was
extended with general neural network features and can now be used to train
standard Artificial Neural Networks (ANNs) and Convolutional Neural Networks
(CNNs)~\cite{Lecun1998} for classification.

While speedups are also observed on the GPU, the proposed library has been
especially optimized for speed on Central Processing Unit (CPU). Although GPUs
are beginning to be the de-facto standard for training deep networks, they are
not always available and some deployments are still targeting existing CPU
implementations. Moreover, inference is generally performed on CPU once the
network has been trained. Therefore, we believe that it remains important to be
able to both train neural networks in reasonable time and achieve fast inference
on CPUs. In this work, we also report successful optimizations on GPU, but we
have to note that advanced parallelization capabilities of GPU where already
well used~\cite{Upadhyaya2013, Lopes2014}, especially for convolutional
networks~\cite{Krizhevsky2012}.

Further to our speedup contributions, a special contribution of this paper is
a comprehensive evaluation against several important state of the art
frameworks. The evaluation is carried on four different models and three data
sets. Comparisons are performed in terms of computation time on both CPU and
GPU, as well as the final accuracy of the trained models.

The rest of this paper is organized as follows. The DLL library is described in
details in Section~\ref{sec:dll}. The experimental evaluation is presented in
Section~\ref{sec:evaluation}. Section~\ref{sec:mnist} is presenting the results
of the experiments on MNIST, Section~\ref{sec:cifar} on CIFAR-10 and
Section~\ref{sec:imagenet} on ImageNet. Finally, conclusions are drawn in
Section~\ref{sec:conclusion}.

\section{DLL: Deep Learning Library}
\label{sec:dll}

Deep Learning Library (DLL) is a Machine Learning framework originally focused
on RBM and CRBM support. It was developed and used in the context of several
research work~\cite{wicht2016deep,wicht2016cpu,wicht2015mixed,wicht2018deep}. It
also has support for various neural network layers and standard backpropagation
techniques. It is written in C++ and its main interface is C++ (example in
Section~\ref{sec:example}). The framework can also be used by describing the
task in a simple descriptor language, to make it easier for researchers.

The framework has complete support for the RBM model~\cite{Smolensky1986}. The
model can be trained using Contrastive Divergence (CD)~\cite{Hinton2006fast}.
The implementation was designed following the model from~\cite{Hinton2012}. It
also supports Deep Belief Network (DBN), pretrained layer by layer and then
fine-tuned using gradient descent.  The RBM supports a wide range of visible
and hidden unit types, such as binary, Gaussian and Rectified Linear Unit
(ReLU)~\cite{Nair2010}.  Support for CRBM is also integrated, following the
model from~\cite{Lee2009}, as well as second version integrating pooling in the
form of Probabilistic Max Pooling.

The framework also supports conventional neural network. As such, ANNs and CNNs
can be trained. Max Pooling and Average Pooling layers are also supported for
CNNs. These networks can be trained with mini-batch gradient descent. The basic
learning options such as momentum and weight decay are supported. The framework
also support advanced techniques such as Dropout~\cite{Hinton2012dropout} and
Batch Normalization~\cite{Ioffe2015}. Finally, optimizers with adaptive learning
rates such as Adagrad~\cite{Duchi2011}, Adadelta~\cite{Zeiler2012} and
Adam~\cite{Kingma2014Adam} are also integrated. The framework also supports
Auto-Encoders~\cite{Bengio2009} and Convolutional
Auto-Encoders~\cite{Masci2011}. They can be trained on noisy input to improve
generalization, a technique known as Denoising Auto-Encoder~\cite{Vincent2008}.


The DLL library is available online\footnote{URL \url{https://github.com/wichtounet/dll}}, free of charge, under
the terms of the MIT open source license. Details of the project as well as
some tutorials are available on the home page.

\subsection{Performance}

The main focus of the library is runtime performance, both for training and for
inference. The DLL framework has been especially optimized for CPU execution.
Nevertheless, the GPU support for most neural networks is also complete.

The implementation uses several techniques to optimize as much as possible the
runtime performance for training and inference. First, all the computations are
performed using single-precision floating point numbers. This leads to a better
data locality and an increased potential for vectorization. On GPU, it would
even be possible to use half-precision, but modern processors do not have
native capabilities for such computations. Another simple optimization is that
all the computations are performed on a batch rather than on one sample at the
time. This has the advantage of leveraging the necessary operations to higher
level computations. Since this is also generally advantageous for the quality
of the training, this is currently the most common way to train a neural
network.

The forward activation of a fully-connected layer for a mini-batch can be
computed with a single matrix-matrix multiplication~\cite{wicht2016cpu}. This
is also possible for the backward pass, by transposing the weight matrix.
Finally, the gradients for the dense layer can also be computed using one
matrix-matrix multiplication. Thus, such a network mainly needs a
good implementation of this operation in order to be fast.

The Basic Linear Algebra Subprograms (BLAS) interface contains a set of small
and highly-optimized kernels for matrix and vector
computation~\cite{Lawson1979}. When using an efficient BLAS library, the
matrix-matrix multiplication operation can be very efficient. Moreover, using a
parallel BLAS library also leads to significantly increased performance for
large layers.  Moreover, although BLAS libraries are highly optimized for very
large matrices, they are not as fast as possible for small matrices. Therefore,
we automatically detect such cases and use custom vectorized kernels for
small matrix multiplications.

Optimization is more complicated for CNNs. Indeed, the dense layers only
account for a small portion of the training time. Convolutional layers use two
forms of convolution. A valid convolution for the forward pass, which shrinks
the representation and a full convolution for the backward pass to expand it.
Every batch of $N$ images is convolved with $K$ kernels.  It is possible to
rearrange an image into columns so that a matrix-matrix multiplication can be
used to compute the valid convolutions of the image and the $K$ kernels at
once~\cite{wicht2016cpu, Ren2015}. This proved to be very efficient for large
images or large kernels. However, when images are small or kernels are very
small, this is not efficient since the rearranging of the input matrix is a
memory intensive operation that will reduce the gains of this reduction.
Therefore, in these cases, we observed that it is more interesting to perform a
real convolution using an highly-optimized implementation.
First, several floating point operations are computed during the same CPU
cycle, using SSE and AVX, a technique known as Single Instruction Multiple Data
(SIMD). Then, to ensure the maximum throughput, the matrices are padded so that
the last dimension is a multiple of the vector size.
Specialized kernels for the most used kernel sizes, such as 3x3 and 5x5, are
also used. Finally, most of the convolutions can be performed in parallel since
there are no dependencies between them. This proved significantly faster than
the reduction to a matrix-matrix multiplication in several configurations.

There are several possible implementations for the full convolution. First, the
operation can be expressed in terms of another operation, the Fast Fourier
Transform (FFT)~\cite{Mathieu2013}. For this, the input image and the kernel
are padded to the size of the output. Then, their transforms  can be computed,
in parallel. The Hadamard product of the input image with the transform of the
kernel is computed.  The inverse transform of this product is the full
convolution. Computing several convolutions of the same image with different
kernels is more efficient since the transform of the input image is only
computed once. In our experiments, we observed that such implementation is very
efficient for large inputs and large kernels, but it is not as interesting for
small configurations. With very small kernels, it is more efficient to pad the
input and the kernels and perform a valid convolution.  Indeed, a full
convolution is equivalent to a valid convolution with some amount of padding.
When the necessary padding is small enough, it becomes significantly faster than
performing the FFTs. The last option is to use an optimized implementation of
the full convolution. However, due to the large number of border cases, this
would only be faster than the implementation as a valid convolution for large
dimensions, in which case the reduction to FFT would be faster.

Unfortunately, there is no one-size-fits-all implementation for all convolution
configurations. Therefore, heuristics are used to select the most suited
implementation for each possible configuration.  These heuristics are based on
the size of the convolution kernels and the size of the batch.

Although most of the time is contained inside the previously mentioned
operations, it is still important to optimize the other operations such as
activation functions and gradient computations. In our implementation, these
operations are vectorized and parallelized to maximize the
processor utilization.

Fortunately, when optimizing for GPU, most of the routines are already
implemented in highly specialized libraries. DLL uses NVIDIA libraries in order
to optimize the most used kernels. NVIDIA CUBLAS is used for the matrix-matrix
multiplications and a few other linear algebra operations and NVIDIA
CUDNN~\cite{Chetlur2014} is used for the machine learning operations such as
convolutions, activation functions and gradients computation.  For other
operations, CUDA kernels have been written to ensure that most of the time is
spent on the GPU. Indeed, when optimizing for GPU, it is most important
to avoid copies between the CPU and GPU. Moreover, most of the kernels are
launched asynchronously, without device synchronization. This significantly
reduces the overhead of CUDA kernel calls.

\subsection{Example}
\label{sec:example}

Figure~\ref{listing:example} shows the code necessary to train a three-layer
fully-connected network on the MNIST data set with the DLL library.
The code starts by loading the MNIST data set in memory. Then, the network is
declared layer by layer. After that, the network training parameters are set
and the training is started. Finally, the accuracy on the test set is computed.

\lstset{language=C++}
\begin{figure}
\begin{lstlisting}[frame=single, basicstyle=\tiny]
using namespace dll;

auto dataset = make_mnist_dataset(batch_size<100>{}, scale_pre<255>{});

using network_type = network_desc<
    network_layers<
        dense_layer<28 * 28, 500, sigmoid>,
        dense_layer<500,     250, sigmoid>,
        dense_layer<250,     10,  softmax>
    >
    , updater<updater_type::MOMENTUM>
    , batch_size<100>
>::network_t;

auto net = std::make_unique<network_type>();

net->learning_rate = 0.1;
net->momentum = 0.9;

net->display();
net->fine_tune(dataset.train(), 50);
net->evaluate(dataset.test());
\end{lstlisting}
\caption{Example from the DLL library to train and evaluate a fully-connected network on the MNIST data set.}
\label{listing:example}
\end{figure}

\section{Experimental Evaluation}
\label{sec:evaluation}

We compared our library against popular frameworks on several experiments. The
time to train each model is compared for each framework, both on CPU and on
GPU. For each experiment, the accuracy of each framework was also computed. It
was shown that all the tested frameworks were all exhibiting comparable
accuracy when trained with the same parameters.


We are underlying here that the goal of these experiments is not to reach state
of the art performance on the tested data sets. Indeed, the models are kept
simple on purpose to allow comparison with a wider range of frameworks.
Moreover, the networks are not always trained for as many epochs as they would
be, if achieving high accuracy was the goal. Finally and very importantly, we
are not aware of the full details of all the frameworks. We did our best to
have similar network architecture and training parameters, but it could be that
some implementation details lead to slightly different training schemes,
explaining differences in terms of execution time.

All the results presented in this chapter have been computed on a Gentoo Linux
machine, with 12 GB of RAM, on an Intel\textsuperscript{\textregistered}
Core\textsuperscript{\texttrademark} i7-2600, running at 3.4 GHz (CPU frequency
scaling has been disabled for the purpose of these tests). Both SSE and AVX
vectorization extensions were enabled on the machine. The BLAS operations are
executed with the Intel\textsuperscript{\textregistered} Math Kernel Library
(MKL), in parallel mode. The GPU used for the benchmarks is a NVIDIA
Geforce\textsuperscript{\textregistered} GTX 960 card.  CUDA 8.0.4.4 and CUDNN
5.0.5 are used. To ensure reproducibility, the source code used for these
experiments is available
online\footnote{\url{https://github.com/wichtounet/frameworks}}.



The following reference frameworks have been selected:

\begin{enumerate}
    \item Caffe~\cite{Jia2014}: Caffe is a high-level Machine Learning framework
        that focuses on speed and expression.  It is developed in C++ and
        is used through a text descriptor language.
        Caffe 1.0 was installed from the sources with GPU and MKL support.
    \item TensorFlow~\cite{Abadi2015}: This is a general low-level framework that allows
        expressing a data flow graph to perform numerical computation. The
        core of the system is written in C++, but the features are
        mostly available through a Python front-end.
        Tensorflow 1.3.1 was installed from the sources with CUDA, CUDNN and MKL support.
    \item Keras~\cite{Chollet2015}: It is a high-level Machine Learning library,
        providing a frontend for either Tensorflow or
        Theano. It is written in Python. It provides a very large number of
        high-level models, easing the development of Machine Learning models. The version 2.0.8
        was installed using the official package with Tensorflow 1.3.1.
    \item Torch~\cite{Collobert2011}: Torch is another low-level Machine Learning
        framework, one of the earliest, started in 2002. It is used through a Lua
        front-end. Although it is a low-level framework, it also contains high-level
        modules for Machine Learning. It was installed from the sources,
        from Git commit 3e9e141 with CUDA and MKL support.
    \item DeepLearning4J~\cite{DeepLearning4j2015}: DeepLearning4J is a deep learning
        framework for Java, written in Java, C and C++. It has a very large set of
        features and focuses on distributed computing. The version 0.9.1 was used,
        from Maven.
\end{enumerate}

The frameworks have been selected based on their popularity and also in order to have
a broad range of programming languages. DLL is used directly from the sources,
with the latest version available at this time (Git commit 2f3c62c).

\section{MNIST}
\label{sec:mnist}

The first experiment is performed on the MNIST data set~\cite{Lecun1998Mnist}.
It is a digit recognition task. The data set is made of 60'000 28x28 grayscale
images for training and 10'000 images for testing. It is a very well-known data
set and has been repeatedly used with most of the existing Machine Learning
algorithms. Although it is considered an easy task, it remains an excellent
problem for comparing frameworks since most of them are using it as
example and make source code available.

\subsection{Fully-Connected Neural Network}
\label{sec:mnist:dense}

The first tested network is a fully-connected three-layer ANN with 500 units in the first
layer, 250 in the second layer and 10 final output units for classification.
The first two layers are using the sigmoid function. The last layer is trained
using a softmax cross entropy loss. The network is trained with mini-batches of
100 images, for 50 epochs, with a learning rate of 0.1 and a momentum of 0.9.
The training accuracy is computed after each epoch and the test accuracy is
computed after the end of the complete training. As an example, the code using
the DLL library is presented in Figure~\ref{listing:example}.

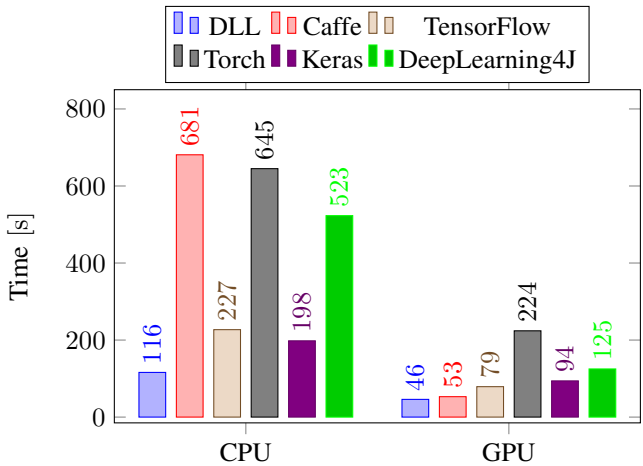
\begin{figure}
\centering
\begin{tikzpicture}
\begin{axis}[
        height=6cm, width=8.5cm,
        symbolic x coords={CPU, GPU},
        xtick=data,
        ylabel=Time,
        y unit=s,
        enlarge x limits = 0.5,
        ymin = -15,
        ymax = 850,
        legend style={at={(0.5,1.25)}, anchor=north, legend columns=3},
        ybar=4pt,
        bar width=10pt,
        nodes near coords,
        every node near coord/.append style={rotate=90, anchor=west},
        outer sep=0pt,
        scaled y ticks=false
    ]
    \addplot coordinates {(CPU,116) (GPU,46)};      
    \addplot coordinates {(CPU,681) (GPU,53)};      
    \addplot coordinates {(CPU,227) (GPU,79)};      
    \addplot coordinates {(CPU,645) (GPU,224)};     
    \addplot coordinates {(CPU,198) (GPU,94)};      
    \addplot coordinates {(CPU,523) (GPU,125)};     
    \legend{DLL,Caffe,TensorFlow,Torch,Keras,DeepLearning4J}
\end{axis}
\end{tikzpicture}
\caption{Training time performance comparison of the frameworks on a Fully-Connected Neural Network experiment, on the MNIST data set, on CPU and on GPU. }
\label{fig:dll:performance:1}
\end{figure}

Figure~\ref{fig:dll:performance:1} presents the runtime performance of each of
the frameworks. In CPU mode, DLL outperforms all the other frameworks, being
around 40\% faster than TensorFlow and Keras. DLL is 4.5 times faster than
DeepLearning4J and  5.5 times faster than Torch and Caffe. On GPU, DLL is the
fastest framework, closely followed by Caffe. DLL is about 40\% faster than
TensorFlow and twice faster than Keras. DeepLearning4J and Torch are
respectively 2.5 and 5 times slower than DLL.



\subsection{Convolutional Neural Network}

The second network, used to solve the same task, is a small CNN with six layers.
The first layer is a convolutional layer using 8 5x5 kernels and followed by
a max pooling layer with a 2x2 kernel. The third is again a convolutional layer
with 8 kernels of dimensions 5x5 and a max pooling layer with 2x2 kernel. The
last layers are fully-connected layers, the first one with 150 units and the
last one with 10 units for classification. The two convolutional layers and the
first fully-connected layer use a sigmoid activation function while the last
layer uses a softmax activation function. The full network is trained in the
same manner as the first network.

\begin{figure}
\centering
\begin{tikzpicture}
\begin{axis}[
        height=6cm, width=8.5cm,
        symbolic x coords={CPU, GPU},
        xtick=data,
        ylabel=Time,
        y unit=s,
        enlarge x limits = 0.5,
        ymin = -50,
        ymax = 3900,
        legend style={at={(0.5,1.25)}, anchor=north, legend columns=3},
        ybar=5pt,
        bar width=10pt,
        nodes near coords,
        every node near coord/.append style={rotate=90, anchor=west},
        outer sep=0pt,
        scaled y ticks=false
    ]
    \addplot coordinates {(CPU,329) (GPU,131)};     
    \addplot coordinates {(CPU,2416) (GPU,564)};    
    \addplot coordinates {(CPU,1021) (GPU,128)};     
    \addplot coordinates {(CPU,1946) (GPU,851)};    
    \addplot coordinates {(CPU,1447) (GPU,185)};    
    \addplot coordinates {(CPU,2800) (GPU,1193)};   
    \legend{DLL,Caffe,TensorFlow,Torch,Keras,DeepLearning4J}
\end{axis}
\end{tikzpicture}
\caption{Training time performance comparison of the frameworks for a CNN, on MNIST, on CPU and on GPU. }
\label{fig:dll:performance:2}
\end{figure}
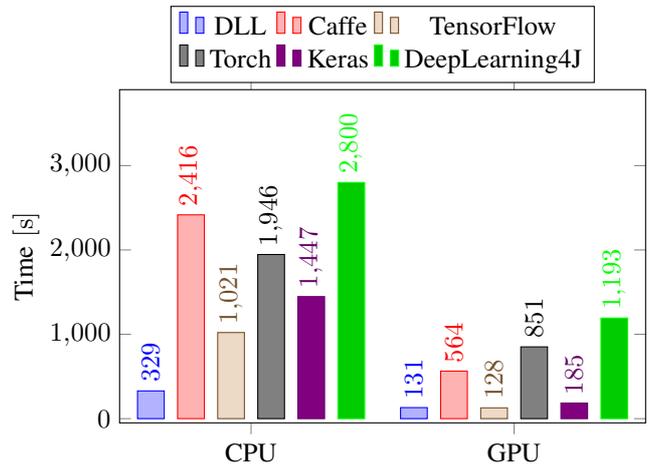

Figure~\ref{fig:dll:performance:2} presents the results obtained on this
experiment. Again, DLL is the fastest framework on CPU, by a significant margin,
three times faster than TensorFlow and almost four times faster than Keras. DLL
is more than 8 times faster than the slowest framework, DeepLearning4J. This
shows the effects of the in-depth CPU optimization of the convolutions. On GPU,
TensorFlow and DLL are the fastest frameworks, about 30\% faster than Keras and
significantly faster than Caffe (4 times), Torch (6.5 times) and DeepLearning4J
(9 times).



\section{CIFAR-10}
\label{sec:cifar}

The second data set that is tested is CIFAR-10~\cite{Krizhevsky2009}, a data set
for object recognition, consisting of 50'000 images for training and 10'000 for
testing, in 10 different classes.  The data set is composed of colour images of
32x32 pixels. The images are made of three colour channels.  This is
a significantly more complicated task than the MNIST task.

A larger CNN is used for this task. The first layer is convolutional with 12 5x5
kernels, followed by a 2x2 max pooling layer. They are followed by another
convolutional layer with 24 3x3 kernels and a 2x2 max pooling layer. A dense
layer with 64 hidden units is then used, followed by a softmax layer with 10
output units. All the layers but the last one are using ReLUs. The network is
trained in a similar manner as the previous networks, with a learning rate of
$0.001$.

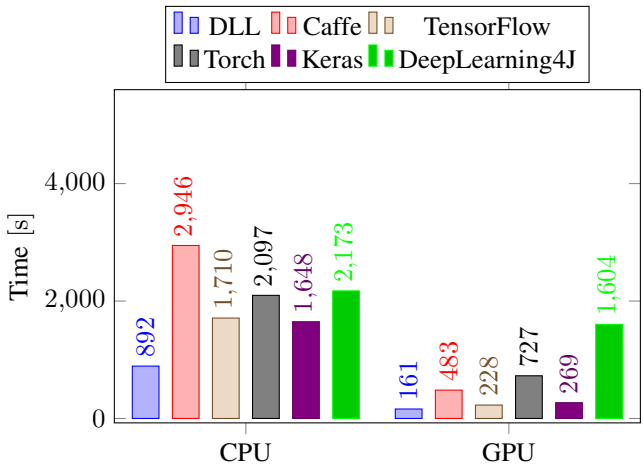
\begin{figure}
\centering
\begin{tikzpicture}
\begin{axis}[
        height=6cm, width=8.5cm,
        symbolic x coords={CPU, GPU},
        xtick=data,
        ylabel=Time,
        y unit=s,
        enlarge x limits = 0.5,
        ymin = -75,
        ymax = 5600,
        legend style={at={(0.5,1.25)}, anchor=north, legend columns=3},
        ybar=5pt,
        bar width=10pt,
        nodes near coords,
        every node near coord/.append style={rotate=90, anchor=west},
        outer sep=0pt,
        scaled y ticks=false
    ]
    \addplot coordinates {(CPU,892) (GPU,161)}; 
    \addplot coordinates {(CPU,2946) (GPU,483)}; 
    \addplot coordinates {(CPU,1710) (GPU,228)}; 
    \addplot coordinates {(CPU,2097) (GPU,727)}; 
    \addplot coordinates {(CPU,1648) (GPU,269)}; 
    \addplot coordinates {(CPU,2173) (GPU,1604)}; 
    \legend{DLL,Caffe,TensorFlow,Torch,Keras,DeepLearning4J}
\end{axis}
\end{tikzpicture}
\caption{Training time performance comparison of the frameworks on the CIFAR-10 task, on CPU and on GPU. }
\label{fig:dll:performance:3}
\end{figure}

In Figure~\ref{fig:dll:performance:3}, the training times for this task are
presented. The speedups are less significant than for the previous CNN.
Nevertheless, DLL still manages to be the fastest framework on CPU. It is about
twice faster than TensorFlow, Keras, DeepLearning4J and Torch and about three
times faster than Caffe.  On GPU, DLL is also the fastest framework on this
experiment, about 30\% faster than TensorFlow and 40\% faster than Keras. It is
three times faster than Caffe and about 4.5 times faster than Torch and ten
times faster than DeepLearning4J.  This network is significantly larger than in
the MNIST experiment. This seems to indicate that most frameworks are more
optimized for larger networks. This also shows that GPU performance is better
when a lot of data is available for computation.



\section{ImageNet}
\label{sec:imagenet}

The last experiment is performed on ImageNet, a very large data set for image
classification. We consider the sub part of the ImageNet Large Scale Visual
Recognition Challenge (ILSVRC) 2012~\cite{ILSVRC15}, there are 50'000
validation images, 100'000 test images, around 1.2 million training images and
1000 categories. As is often the case with this data set, all the images have
     been resized to 256x256 images.


Since the number of images is very large, the entire data set cannot be kept in
memory. Therefore, the images are loaded from the disk for each epoch. Contrary
to the previous tests,
only Caffe provides an official, up-to-date, code for this data set. The
DeepLearning4J reader was based on existing official reader for structures
similar to ImageNet. For Keras, TensorFlow and Torch, a simple data reader has
been written with the image loading tools available in each framework.

The network is significantly larger than the previous networks. It is made of
five convolutional layers, with 16 3x3 kernels for the first two layers and 32
3x3 kernels for the next three layers. Each of these layers is followed by
a ReLU activation function and a 2x2 max pooling layer. All the convolutional
layers are using zero-padding so that their output is the same size as their
input The last two layers are a dense layer with 2048 hidden units, with a ReLU
function and a dense layer with 1000 outputs and a softmax activation function.
The training is different than for the other data sets.  The full network is
only trained for five epochs with each framework. The networks are trained using
a batch size of 128. However, Torch and DeepLearning4J models were trained with
a batch size of 64, respectively 16, samples. Indeed, both of these frameworks
needed more than 12GB of RAM to train with a batch size of 128 images. This may
lead to some small degradation of the performance for those two frameworks.

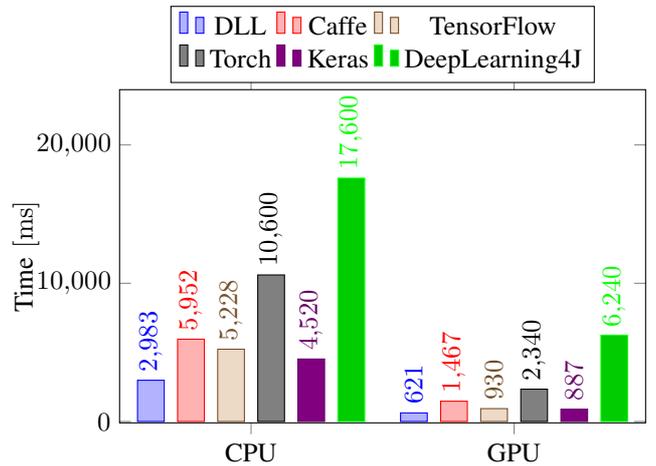
\begin{figure}
\centering
\begin{tikzpicture}
\begin{axis}[
        height=6cm, width=8.5cm,
        symbolic x coords={CPU, GPU},
        xtick=data,
        ylabel=Time,
        y unit=ms,
        enlarge x limits = 0.5,
        ymin = -100,
        ymax = 24000,
        legend style={at={(0.5,1.25)}, anchor=north, legend columns=3},
        ybar=5pt,
        bar width=10pt,
        nodes near coords,
        every node near coord/.append style={rotate=90, anchor=west},
        outer sep=0pt,
        scaled y ticks=false
    ]
    \addplot coordinates {(CPU,2983) (GPU,621)}; 
    \addplot coordinates {(CPU,5952) (GPU,1467)}; 
    \addplot coordinates {(CPU,5228) (GPU,930)}; 
    \addplot coordinates {(CPU,10600) (GPU,2340)}; 
    \addplot coordinates {(CPU,4520) (GPU,887)}; 
    \addplot coordinates {(CPU,17600) (GPU,6240)}; 
    \legend{DLL,Caffe,TensorFlow,Torch,Keras,DeepLearning4J}
\end{axis}
\end{tikzpicture}
\caption{Training time performance comparison of the frameworks on the ImageNet task, on CPU and on GPU. The time is the average time necessary for the training of one batch of 128 elements.}
\label{fig:dll:performance:4}
\end{figure}

For the sake of comparison, the average time to train one batch of samples is
used as results. For Torch and DeepLearning4J, the results are the times for
several batches, to make up for 128 samples. These results are presented in
Figure~\ref{fig:dll:performance:4}. DLL shows to be again the fastest framework
on CPU for training this large model, 35\% faster than Keras, about 45\% faster
than TensorFlow and twice faster than Caffe. Torch is already more than 3 times
slower than DLL and DeepLearning4J around 6 times slower. On GPU, DLL is, also,
the fastest framework.  Comparisons with Keras and TensorFlow show that most of
the difference comes from the poor performance of reading the ImageNet data from
the Python code. Once this is taken into account, the three frameworks have
comparable performance. DLL is more than twice faster than Caffe and almost four
times faster than Torch and almost 10 times faster than DeepLearning4J.



\section{Conclusion and Future Work}
\label{sec:conclusion}

For all the experiments and the different neural networks models that were
tested, the DLL framework has shown to be the fastest gradient descent based
framework for training the model when using CPU and GPU. For each test, the
accuracies of the models trained with DLL are similar to the models trained
by the other five Machine Learning frameworks.

The speedups provided by the framework on CPU mode are especially important for
convolutional layers for which advanced optimization was performed. The
framework was especially optimized for small convolutions, but is still able to
bring significant speedups for large images such as the images from the ImageNet
data set. Moreover, while some frameworks are mostly optimized for the
convolutional and fully-connected parts of the computation, every part of the
training in the DLL framework was tuned.


While the library is highly optimized for small images, its performance should
be improved further for large images. As potential improvement we believe that
different optimized kernels should be taken not only depending on the size of
the kernel but also on the size of the image. Also, a few DLL routines are not
optimized enough for GPU, such as Dropout and Batch Normalization. Future work
could also include better support for Recurrent Neural Networks (RNNs), which
would be a great advantage for the library. Finally, the library has currently
been optimized only on few machines and especially consumer grade processors and
graphics cards. It would be greatly beneficial to take advantage of more threads
or advanced vectorization capabilities such as those provided by the latest
Intel\textsuperscript{\textregistered} Xeon processors or more recent and
powerful NVIDIA graphics cards.


\appendices

\bibliographystyle{../IEEEconf_ICPR18/IEEEtran}
\bibliography{bibliography}

\begin{thebibliography}{10}
\providecommand{\url}[1]{#1}
\csname url@samestyle\endcsname
\providecommand{\newblock}{\relax}
\providecommand{\bibinfo}[2]{#2}
\providecommand{\BIBentrySTDinterwordspacing}{\spaceskip=0pt\relax}
\providecommand{\BIBentryALTinterwordstretchfactor}{4}
\providecommand{\BIBentryALTinterwordspacing}{\spaceskip=\fontdimen2\font plus
\BIBentryALTinterwordstretchfactor\fontdimen3\font minus
  \fontdimen4\font\relax}
\providecommand{\BIBforeignlanguage}[2]{{%
\expandafter\ifx\csname l@#1\endcsname\relax
\typeout{** WARNING: IEEEtran.bst: No hyphenation pattern has been}%
\typeout{** loaded for the language `#1'. Using the pattern for}%
\typeout{** the default language instead.}%
\else
\language=\csname l@#1\endcsname
\fi
#2}}
\providecommand{\BIBdecl}{\relax}
\BIBdecl

\bibitem{Szegedy2015}
C.~Szegedy, W.~Liu, Y.~Jia, P.~Sermanet, S.~Reed, D.~Anguelov, D.~Erhan,
  V.~Vanhoucke, and A.~Rabinovich, ``Going deeper with convolutions,'' in
  \emph{Proceedings of the IEEE Conference on Computer Vision and Pattern
  Recognition}, 2015, pp. 1--9.

\bibitem{Karpathy2015}
A.~Karpathy and L.~Fei-Fei, ``Deep visual-semantic alignments for generating
  image descriptions,'' in \emph{Proceedings of the IEEE Conference on Computer
  Vision and Pattern Recognition}, 2015, pp. 3128--3137.

\bibitem{Cheng2015}
Z.~Cheng, Q.~Yang, and B.~Sheng, ``Deep colorization,'' in \emph{Proceedings of
  the IEEE International Conference on Computer Vision}, 2015, pp. 415--423.

\bibitem{Goodfellow2014}
I.~Goodfellow, J.~Pouget-Abadie, M.~Mirza, B.~Xu, D.~Warde-Farley, S.~Ozair,
  A.~Courville, and Y.~Bengio, ``Generative adversarial nets,'' in
  \emph{Advances in neural information processing systems}, 2014, pp.
  2672--2680.

\bibitem{Smolensky1986}
P.~Smolensky, ``Information processing in dynamical systems: Foundations of
  harmony theory,'' 1986.

\bibitem{Lee2009}
H.~Lee, R.~Grosse, R.~Ranganath, and A.~Y. Ng, ``Convolutional deep belief
  networks for scalable unsupervised learning of hierarchical
  representations,'' in \emph{Proceedings of the 26th Annual International
  Conference on Machine Learning}.\hskip 1em plus 0.5em minus 0.4em\relax ACM,
  2009, pp. 609--616.

\bibitem{Lecun1998}
Y.~LeCun, L.~Bottou, Y.~Bengio, and P.~Haffner, ``Gradient-based learning
  applied to document recognition,'' \emph{Proceedings of the IEEE}, vol.~86,
  no.~11, pp. 2278--2324, 1998.

\bibitem{Upadhyaya2013}
S.~R. Upadhyaya, ``Parallel approaches to machine learning: A comprehensive
  survey,'' \emph{Journal of Parallel and Distributed Computing}, vol.~73,
  no.~3, pp. 284--292, 2013.

\bibitem{Lopes2014}
N.~Lopes and B.~Ribeiro, ``Towards adaptive learning with improved convergence
  of {Deep Belief Networks} on {Graphics Processing Units},'' \emph{Pattern
  Recognition}, vol.~47, no.~1, pp. 114--127, 2014.

\bibitem{Krizhevsky2012}
A.~Krizhevsky, I.~Sutskever, and G.~E. Hinton, ``Imagenet classification with
  deep convolutional neural networks,'' in \emph{Advances in neural information
  processing systems}, 2012, pp. 1097--1105.

\bibitem{wicht2016deep}
B.~Wicht, A.~Fischer, and J.~Hennebert, ``Deep learning features for
  handwritten keyword spotting,'' in \emph{Pattern Recognition (ICPR), 2016
  23rd International Conference on}.\hskip 1em plus 0.5em minus 0.4em\relax
  IEEE, 2016, pp. 3434--3439.

\bibitem{wicht2016cpu}
------, ``On cpu performance optimization of restricted boltzmann machine and
  convolutional rbm,'' in \emph{IAPR Workshop on Artificial Neural Networks in
  Pattern Recognition}.\hskip 1em plus 0.5em minus 0.4em\relax Springer
  International Publishing, 2016, pp. 163--174.

\bibitem{wicht2015mixed}
B.~Wicht and J.~Hennebert, ``Mixed handwritten and printed digit recognition in
  sudoku with convolutional deep belief network,'' in \emph{Document Analysis
  and Recognition (ICDAR), 2015 13th International Conference on}.\hskip 1em
  plus 0.5em minus 0.4em\relax IEEE, 2015, pp. 861--865.

\bibitem{wicht2018deep}
B.~Wicht, ``Deep learning features for image processing,'' Ph.D. dissertation,
  University of Fribourg, 2018.

\bibitem{Hinton2006fast}
G.~E. Hinton, S.~Osindero, and Y.-W. Teh, ``A fast learning algorithm for deep
  belief nets,'' \emph{Neural computation}, vol.~18, no.~7, pp. 1527--1554,
  2006.

\bibitem{Hinton2012}
G.~E. Hinton, ``A practical guide to training restricted boltzmann machines.''
  in \emph{Neural Networks: Tricks of the Trade (2nd ed.)}, ser. Lecture Notes
  in Computer Science.\hskip 1em plus 0.5em minus 0.4em\relax Springer, 2012,
  vol. 7700, pp. 599--619.

\bibitem{Nair2010}
V.~Nair and G.~E. Hinton, ``{Rectified Linear Units} improve {Restricted
  Boltzmann Machines},'' in \emph{Proceedings of the Int. Conf. on Machine
  Learning}, 2010, pp. 807--814.

\bibitem{Hinton2012dropout}
G.~E. Hinton, N.~Srivastava, A.~Krizhevsky, I.~Sutskever, and R.~R.
  Salakhutdinov, ``Improving neural networks by preventing co-adaptation of
  feature detectors,'' \emph{arXiv preprint arXiv:1207.0580}, 2012.

\bibitem{Ioffe2015}
S.~Ioffe and C.~Szegedy, ``Batch normalization: Accelerating deep network
  training by reducing internal covariate shift,'' \emph{arXiv preprint
  arXiv:1502.03167}, 2015.

\bibitem{Duchi2011}
J.~Duchi, E.~Hazan, and Y.~Singer, ``Adaptive subgradient methods for online
  learning and stochastic optimization,'' \emph{Journal of Machine Learning
  Research}, vol.~12, no. Jul, pp. 2121--2159, 2011.

\bibitem{Zeiler2012}
M.~D. Zeiler, ``Adadelta: an adaptive learning rate method,'' \emph{arXiv
  preprint arXiv:1212.5701}, 2012.

\bibitem{Kingma2014Adam}
D.~Kingma and J.~Ba, ``Adam: A method for stochastic optimization,''
  \emph{arXiv preprint arXiv:1412.6980}, 2014.

\bibitem{Bengio2009}
Y.~Bengio, ``Learning deep architectures for ai,'' \emph{Foundations and
  trends{\textregistered} in Machine Learning}, vol.~2, no.~1, pp. 1--127,
  2009.

\bibitem{Masci2011}
J.~Masci, U.~Meier, D.~Cire{\c{s}}an, and J.~Schmidhuber, \emph{Stacked
  Convolutional Auto-Encoders for Hierarchical Feature Extraction}.\hskip 1em
  plus 0.5em minus 0.4em\relax Springer Berlin Heidelberg, 2011, pp. 52--59.

\bibitem{Vincent2008}
P.~Vincent, H.~Larochelle, Y.~Bengio, and P.-A. Manzagol, ``Extracting and
  composing robust features with denoising autoencoders,'' in \emph{Proceedings
  of the 25th international conference on Machine learning}.\hskip 1em plus
  0.5em minus 0.4em\relax ACM, 2008, pp. 1096--1103.

\bibitem{Lawson1979}
C.~L. Lawson, R.~J. Hanson, D.~R. Kincaid, and F.~T. Krogh, ``Basic linear
  algebra subprograms for fortran usage,'' \emph{ACM Transactions on
  Mathematical Software (TOMS)}, vol.~5, no.~3, pp. 308--323, 1979.

\bibitem{Ren2015}
J.~S. Ren and L.~Xu, ``On vectorization of deep convolutional neural networks
  for vision tasks,'' \emph{arXiv preprint arXiv:1501.07338}, 2015.

\bibitem{Mathieu2013}
M.~Mathieu, M.~Henaff, and Y.~LeCun, ``Fast training of convolutional networks
  through ffts,'' \emph{arXiv preprint arXiv:1312.5851}, 2013.

\bibitem{Chetlur2014}
\BIBentryALTinterwordspacing
S.~Chetlur and al., ``cudnn: Efficient primitives for deep learning,''
  \emph{CoRR}, vol. abs/1410.0759, 2014. [Online]. Available:
  \url{http://arxiv.org/abs/1410.0759}
\BIBentrySTDinterwordspacing

\bibitem{Jia2014}
Y.~Jia, E.~Shelhamer, J.~Donahue, S.~Karayev, J.~Long, R.~Girshick,
  S.~Guadarrama, and T.~Darrell, ``Caffe: Convolutional architecture for fast
  feature embedding,'' \emph{arXiv preprint arXiv:1408.5093}, 2014.

\bibitem{Abadi2015}
\BIBentryALTinterwordspacing
M.~Abadi and al., ``{TensorFlow}: Large-scale machine learning on heterogeneous
  systems,'' 2015, software available from tensorflow.org. [Online]. Available:
  \url{http://tensorflow.org/}
\BIBentrySTDinterwordspacing

\bibitem{Chollet2015}
F.~Chollet, ``keras,'' \url{https://github.com/fchollet/keras}, 2015.

\bibitem{Collobert2011}
R.~Collobert, K.~Kavukcuoglu, and C.~Farabet, ``Torch7: A matlab-like
  environment for machine learning,'' in \emph{BigLearn, NIPS Workshop}, 2011.

\bibitem{DeepLearning4j2015}
\BIBentryALTinterwordspacing
D.~D. Team, ``Deeplearning4j: Open-source distributed deep learning for the
  jvm,'' 2015. [Online]. Available: \url{http://deeplearning4j.org}
\BIBentrySTDinterwordspacing

\bibitem{Lecun1998Mnist}
Y.~LeCun, C.~Cortes, and C.~J.~C. Burges, ``The mnist database of handwritten
  digits,'' 1998.

\bibitem{Krizhevsky2009}
A.~Krizhevsky and G.~E. Hinton, ``Learning multiple layers of features from
  tiny images,'' 2009.

\bibitem{ILSVRC15}
O.~Russakovsky and al., ``{ImageNet Large Scale Visual Recognition
  Challenge},'' \emph{International Journal of Computer Vision (IJCV)}, vol.
  115, no.~3, pp. 211--252, 2015.

\end{thebibliography}

\end{document}